\documentclass[conference,a4paper]{IEEEtran}
\IEEEoverridecommandlockouts

\usepackage[hidelinks]{hyperref}
\usepackage[cmex10]{amsmath}
\usepackage{amssymb,amsfonts}
\usepackage{dblfloatfix}

\usepackage[ruled,vlined]{algorithm2e}
\usepackage{graphicx}
\graphicspath{{Figures/PDF/}{Figures/PNG/}}

\usepackage{booktabs}
\usepackage{siunitx}
\usepackage[numbers,compress]{natbib}
\usepackage{texnames}
\usepackage{bm,bbm}
\usepackage{orcidlink}
\usepackage{lettrine}
 
\begin{document}
\title{SpecTM: Spectral Targeted Masking for Trustworthy Foundation Models}
\author{
    \IEEEauthorblockN{
        Syed Usama Imtiaz\orcidlink{0000-0002-1229-0100},
        Mitra Nasr Azadani\orcidlink{0009-0001-3508-545X}, and
        Nasrin Alamdari\orcidlink{0000-0001-6595-6669}
    }
    \IEEEauthorblockA{
        \textit{Department of Civil and Environmental Engineering
        Florida State University, Tallahassee, FL, USA}\\
        \{si22j, mn22, nalamdari\}@fsu.edu
    }
   \thanks{Copyright 2026 IEEE. Published in the 2026 IEEE International Geoscience and Remote Sensing Symposium (IGARSS 2026), scheduled for 9 - 14 August 2026 in Washington, D.C.. Personal use of this material is permitted. However, permission to reprint/republish this material for advertising or promotional purposes or for creating new collective works for resale or redistribution to servers or lists, or to reuse any copyrighted component of this work in other works, must be obtained from the IEEE. Contact: Manager, Copyrights and Permissions / IEEE Service Center / 445 Hoes Lane / P.O. Box 1331 / Piscataway, NJ 08855-1331, USA. Telephone: + Intl. 908-562-3966.}
}
\maketitle
 
 
\begin{abstract} 

Foundation models are now increasingly being developed for Earth observation (EO), yet they often rely on stochastic masking that do not explicitly enforce physics constraints; a critical trustworthiness limitation, in particular for predictive models that guide public health decisions. In this work, we propose \textit{SpecTM (Spectral Targeted Masking)}, a physics-informed masking design that encourages the reconstruction of targeted bands from cross-spectral context during pretraining. To achieve this, we developed an adaptable multi-task (band reconstruction, bio-optical index inference, and 8-day-ahead temporal prediction) self-supervised learning (SSL) framework that encodes spectrally intrinsic representations via joint optimization, and evaluated it on a downstream \textit{microcystin} concentration regression model using NASA PACE hyperspectral imagery over Lake Erie. SpecTM achieves $R^2{=}0.695$ (current week) and $R^2{=}0.620$ (8-day-ahead) predictions surpassing all baseline models by ($+$34\% (0.51 Ridge) and $+$99\% (SVR 0.31)) respectively. Our ablation experiments show targeted masking improves predictions by $+\!0.037$ $R^2$ over random masking. Furthermore, it outperforms strong baselines with $2.2\times$ superior label efficiency under extreme scarcity. SpecTM enables physics-informed representation learning across EO domains and improves the interpretability of foundation models.

\end{abstract}


\section{Introduction}
\label{sec:intro}

\lettrine{M}{asked image modeling} (MIM) based foundation models~\cite{sun2023ringmo} learn representations by randomly masking a high proportion of the input image (processed as patch tokens) and reconstructing the masked regions from the visible context. Yet, random masking provides no explicit physical constraint on \textit{what} the model learns, and can rely on spurious correlations present in the training distribution. In Earth observation (EO) models, particularly for applications that inform public health decisions, this creates a trustworthiness gap. Here, trustworthiness refers to developing models that are  interpretable on established domain knowledge learning. Satellite remote sensing (SRS) utilizing hyperspectral imagery, such as by NASA's recently launched Plankton, Aerosol, Cloud, Ocean Ecosystem (PACE) mission~\cite{werdell2019pace}, provides unprecedented spectral observational details with high-fidelity that can help close this gap. While natural RGB images mainly learn from spatial patterns (shapes and textures) with limited spectral information, hyperspectral data encode physics(absorption/scattering) of constituents at specific wavelengths. For instance, phycocyanin absorption near $\lambda \ 620\,\mathrm{nm}$ is a marker of cyanobacterial biomass and Chlorophyll-\textit{a} absorption near $\lambda \ 665\,\mathrm{nm}$ indicates phytoplankton (photosynthetic pigment) abundance~\cite{mishra2013phycocyanin,lyu2013phycocyanin,ho2017light}. The targeted masking of these pigment-sensitive bands would encourage the model to learn established bio-optical relationships rather than stochastic patterns. Current methods, such as SpectralGPT~\cite{hong2024spectralgpt} apply 3D spatial spectral masking using a high masking ratio (90\%), SatMAE~\cite{cong2022satmae} temporally masks spatial patches, and TerraMAE~\cite{faruk2025terrammae} groups bands using statistical reflectance correlation. Yet these methods share a critical limitation: spectral bands are masked either uniformly at random or via statistical groupings. These methods implicitly assume that all wavelengths are equally informative. We propose \textit{SpecTM (Spectral Targeted Masking)}, a targeted masking design that masks pigment-sensitive bands to learn cross-band spectral dependencies via reconstruction. We validate our approach on the downstream task of \textit{microcystin} concentration prediction using NASA PACE hyperspectral imagery over Lake Erie, a region with annual economic impacts estimated at over \$70 million. Our contributions: \begin{enumerate}
\item We posit that embedding domain knowledge for masking strategy bridges the trustworthiness gap in spectral foundation models (Section~\ref{sec:method}).
    
\item We introduced an adaptable self-supervised learning (SSL) framework for hyperspectral imagery that encodes spectrally intrinsic representations via multi-task joint optimization: band reconstruction, bio-optical index inference, and 8-day-ahead temporal prediction (Section~\ref{sec:pretraining}).
    
\item We demonstrate direct hyperspectral-to-\textit{microcystin} regression using lab measurements as ground truth, without relying on proxies (e.g., chlorophyll-\textit{a}) or threshold-based classification; to our knowledge, this is the first \textit{direct 8-day-Ahead microcystin} prediction from hyperspectral imagery. (Section~\ref{sec:experiments}).
\end{enumerate}

\section{Related Work}
\label{sec:related}

In the realm of SSL, MIM leverages unlabeled datasets through pretext tasks to learn improved representations. Masked Autoencoders (MAEs) ~\cite{he2022mae} have catalyzed this paradigm shift in representation learning for Earth observation, such as SpectralGPT~\cite{hong2024spectralgpt}, a large (600M+) spectral remote-sensing foundation model trained on multispectral (Sentinel-2) data. However, architectural adaptations have recently emerged to enhance the efficacy of learning from satellite data. For instance, SatMAE~\cite{cong2022satmae}introduces temporal positional embeddings, while SS-MAE~\cite{lin2023ssmae}, a dual-branch architecture that learns spatial and spectral representations in parallel. MIM was further refined in TerraMAE~\cite{faruk2025terrammae}by incorporating spectral-fidelity losses that better preserve inter-band correlations. Despite these architectural advancements, the masking strategy, particularly physics-informed spectral band masking, remains underexplored; such physically relevant band learning holds immense potential to improve both performance and the trustworthiness of foundation models.
Algal bloom monitoring ~\cite{alamdari2026algal}, now increasingly satellite-based, exemplifies the necessity for physics-informed spectral analysis for downstream prediction tasks that inform public health decisions. In this regard, the predictive strength of the Cyanobacteria Index, and the phycocyanin algorithms are derived precisely from targeting wavelengths with known biogeochemical significance that are directly encoded within the algorithm formulations ~\cite{mishra2021microcystin}. In addition, artificial intelligence (AI) is increasingly being applied to environmental ~\cite{rabby2026mlwq,nuriddinov2026high} and bacterial modeling ~\cite{salou2026ecoli,azadani2026picsrl}. Predictive modeling for bloom presence has achieved strong accuracy~\cite{hill2020habnet, acunaalonso2025dlhab}, and our prior work using contrastive SSL improved detection from satellite imagery~\cite{imtiaz2025simclr}. Bloom toxicity cannot be directly observed from satellite imagery, and continuous toxin concentration regression remains challenging. \textit{Microcystin} cyanotoxin concentrations exhibit a nonlinear and temporally variable dependence on bloom biomass~\cite{stumpf2016cyanotoxin}, and this challenge is further compounded by the scarcity of labeled observations, and inherent inland hydrology ~\cite{azadani2025impoundment}. While SSL offers a potential solution to label scarcity, it doesn’t impose physics-based spectral constraints. These constraints underpin the trustworthiness of bio-optical algorithms.

\section{Method}
\subsection{\textbf{Targeted Masking Formalization}} 

\label{sec:method}  
Let $\mathbf{x} \in \mathbb{R}^B$ represent a $B$-band per-pixel  hyperspectral reflectance spectrum and let $\mathcal{D} \subset  \{1, \ldots, B\}$ denote the set of \textit{diagnostic band indices}  identified from domain knowledge. Targeted masking deterministically  defines the mask: \begin{equation} m_b = \mathbf{1}[b \in \mathcal{D}], \end{equation} where $m_b \in \{0, 1\}$ indicates whether band $b$ is masked. All  diagnostic bands are masked while context bands remain fully visible.  For cyanobacteria monitoring, we define $\mathcal{D}$ to include  three bio-optically significant spectral regions: phycocyanin  absorption (615--640\,nm), chlorophyll-\textit{a} red absorption  (660--680\,nm), and the red/NIR transition region (695--720\,nm),  spanning 28 of 122 PACE OCI bands. Masked bands are zeroed before  spectral tokenization (rather than replaced with learnable mask  vectors) so that masked-band values do not contribute to aggregated  spectral tokens when multiple bands are combined. The selection  of $\mathcal{D}$ follows established phytoplankton  optics~\cite{wynne2010lakeerie}.

\subsection{\textbf{Architecture}}  
\begin{figure*}[!t]
\centering
\includegraphics[width=0.88\textwidth]{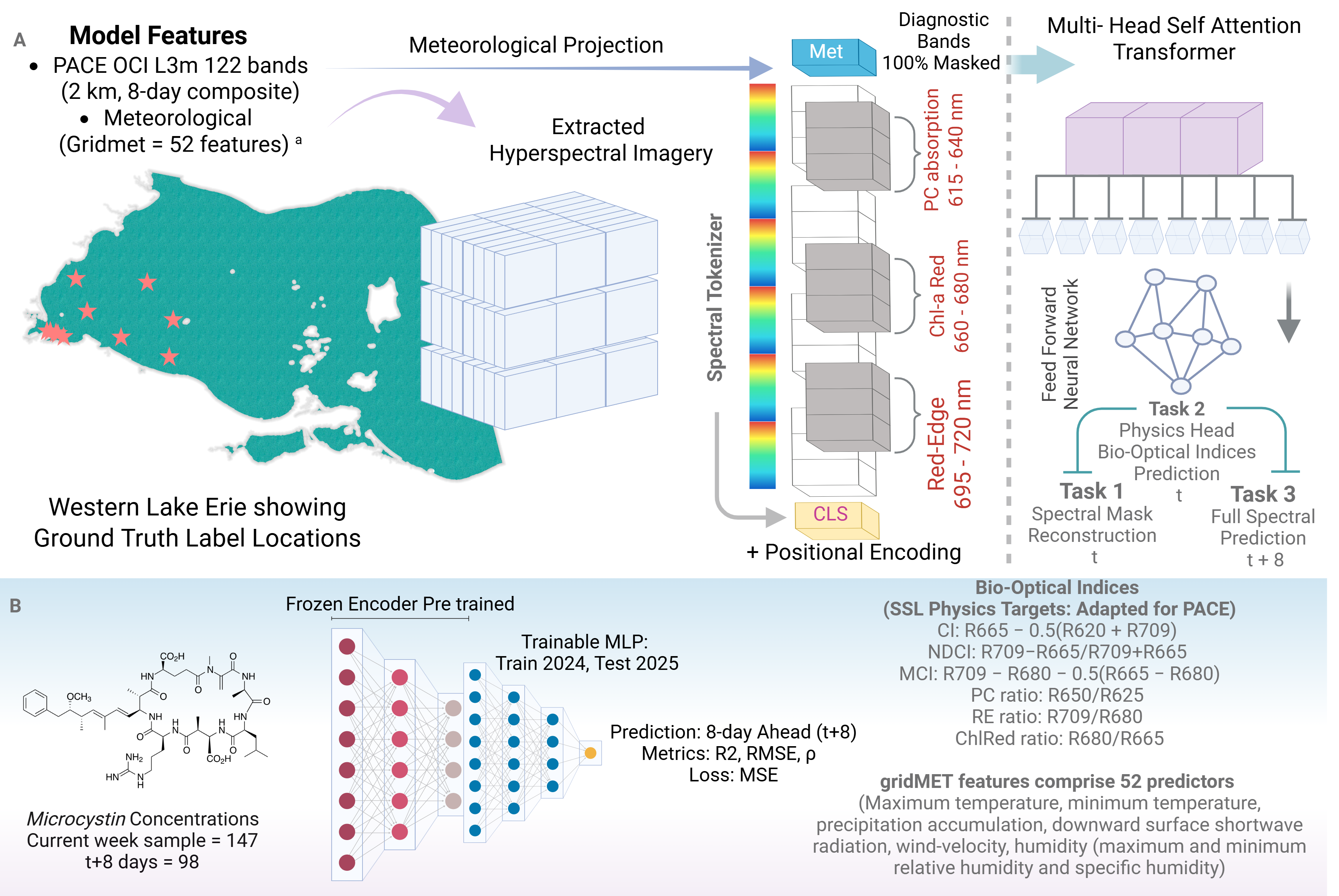}
\vspace{-3pt}
\caption{(A) PACE OCI Level-3 hyperspectral imagery (122 bands, 2 km, 8-day composite) is combined with meteorological predictors (GridMET; 52 features). Spectral tokens are constructed and augmented with positional encoding, while key diagnostic bands (e.g., PC absorption, chlorophyll-a red, and red-edge regions) are masked to enforce physics-guided learning. A multi-head self-attention transformer processes the inputs and is jointly trained on three tasks. (B) The pretrained encoder is frozen and coupled with a trainable MLP head to predict microcystin concentrations 8 days ahead, evaluated using R², RMSE, and correlation $\rho$ and lists Bio-optical indices and meteorological variables used.}
\label{fig:igrass_spectm_2026}
\end{figure*}

We employ a Vision Transformer encoder with 6 layers, 8 attention heads,  
embedding dimension $d{=}256$, and feedforward dimension 1024 ($\sim$6M  
parameters). The 122 PACE OCI bands ($\sim$345--2260\,nm) are tokenized into  
12 contiguous spectral tokens at predefined bio-optically meaningful boundaries (Figure~\ref{fig:igrass_spectm_2026}). A 2-layer transformer decoder reconstructs band-level outputs from the encoded token representations. A learnable classification (CLS) token is prepended to the spectral token sequence; its final representation serves as the global  
encoding for downstream prediction heads. Meteorological features from the gridMET dataset (52 total: 10 base variables 
at $t$, $t{-}8$, $t{-}16$, $t{-}32$ days, plus 12 derived features) are 
incorporated as a separate meteorology token, conditioning spectral 
reconstruction on environmental drivers.
\setlength{\abovecaptionskip}{0.25pt}
\setlength{\belowcaptionskip}{0.25pt}
\subsection{\textbf{Multi-task Pretraining}}
\label{sec:pretraining}

We jointly optimize three self-supervised objectives that learned 
complementary aspects of spectral-temporal structure. \textit{Spectral 
reconstruction} ($\mathcal{L}_{\text{recon}}$) minimizes MSE between 
predicted and true reflectance at masked band positions only, forcing 
the encoder to infer diagnostic bands from spectral context. 
\textit{Physics index prediction} ($\mathcal{L}_{\text{phys}}$) regresses 
six bio-optical indices (Figure~\ref{fig:igrass_spectm_2026}) from the CLS token 
embedding; since defining bands are deterministically masked, success 
requires learning spectral covariance structure rather than direct 
computation. \textit{Temporal forecasting} ($\mathcal{L}_{\text{temp}}$) 
predicts the full spectrum at $t{+}8$ days (8-day composite cadence), 
encouraging the encoder to capture bloom temporal dynamics. The combined 
loss is:
\begin{equation}
\mathcal{L}_{\text{SSL}} = \lambda_1 \mathcal{L}_{\text{recon}} + \lambda_2 \mathcal{L}_{\text{phys}} + \lambda_3 \mathcal{L}_{\text{temp}}
\end{equation}
with weights $\lambda_1{=}1.0$, $\lambda_2{=}0.5$, $\lambda_3{=}0.3$ 
(selected via grid search on SSL validation loss). For SSL pretraining, we train for 100 epochs with AdamW (lr $10^{-4}$, weight decay 0.01), batch size 256, and cosine schedule with 5-epoch 
warmup ($\sim$4 hours on NVIDIA A100). Pretraining quality is measured 
by masked-position Pearson $r$ and physics-index $R^2$.

\subsection{\textbf{Fine-Tuning and Evaluation}}  
For downstream \textit{microcystin} prediction, we freeze the pretrained encoder 
to prevent overfitting on the small labeled dataset and train a 2-layer 
MLP head (hidden dimensions 128, 64; dropout 0.3) using MSE loss on 
the log-transformed target $y = \log(1 + \text{MC})$.For fine-tuning, 
we use AdamW (lr $10^{-3}$), up to 100 epochs, and early stopping 
(patience 20). Current week composite ($n{=}147$) uses Leave-One-Group-Out CV by 
8-day composite period; 8-day-ahead prediction ($n{=}98$ consecutive pairs) 
2024 $\rightarrow$ 2025 temporal split. We report $R^2$, RMSE, and 
Spearman $\rho$.


\section{Experiments}
We selected Western Lake Erie for its recurrent cyanobacterial blooms and intensive monitoring coverage, enabling systematic pairing of hyperspectral observations with in-situ toxin measurements. Figure~\ref{fig:igrass_spectm_2026} summarizes our experimental workflow.  For self-supervised pretraining, we utilized NASA PACE OCI Level-3 mapped  8-day composites (2\,km resolution) acquired between April 2024 and  August 2025, yielding 71,320 spectral-meteorological pairs. Preprocessing  follows our previous validation protocol~\cite{imtiaz2025simclr}. We  obtained \textit{microcystin} concentrations from NOAA GLERL weekly  sampling with strict alignment criteria ($\leq$2\,km, $\pm$4\,days),  returning 147 matched samples (0.10--10.70\,$\mu$g/L) and 98 temporally  paired observations for 8-day-ahead prediction.  We benchmarked against 26,208 baseline configurations spanning seven  algorithms and 78 feature combinations, using a strict temporal split  (Train: 2024, $n{=}112$; Test: 2025, $n{=}35$). Ablation analyses isolate  contributions of targeted masking and quantify label efficiency  (5--100\% training fractions). All baselines received identical input features: 122 spectral bands concatenated with 52 meteorological features, ensuring fair comparison.


\begin{figure}[t]
\centering
\includegraphics[width=\columnwidth]{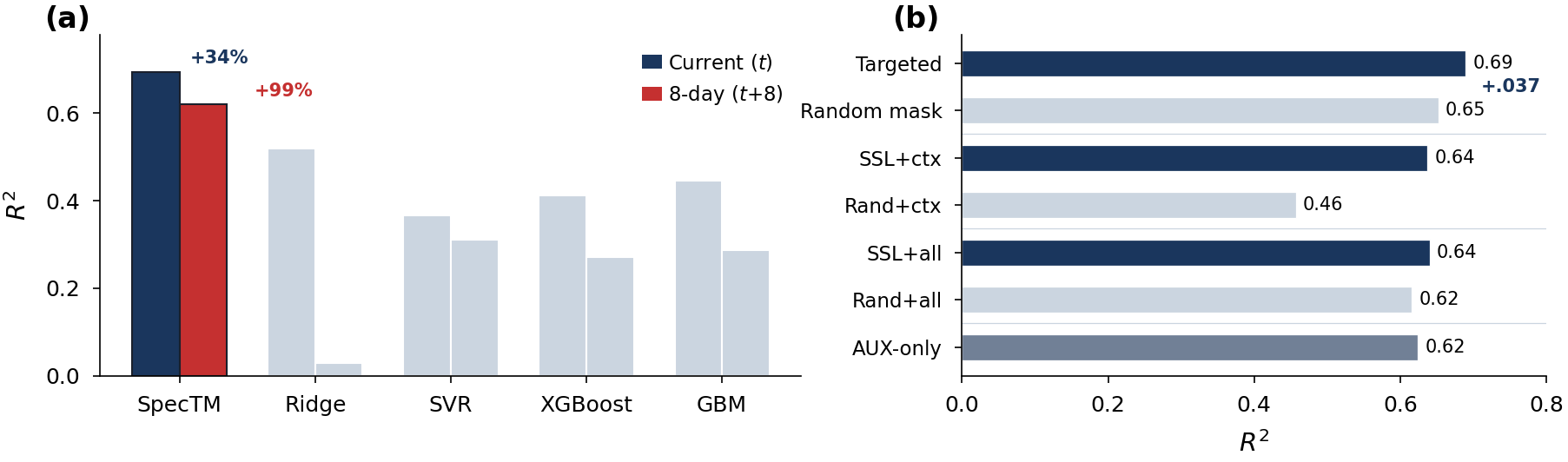}
\vspace{-7mm}
\caption{Experimental results. (a) SpecTM outperforms all baselines for both current-week (+34\%) and 8-day-ahead (+99\%) microcystin prediction. (b) Ablation analysis: targeted masking improves over random by +0.037 $R^2$; SSL pretraining provides +0.18 $R^2$ gain over random initialization.}
\label{fig:combined_v3}
\end{figure}

\subsection{\textbf{SSL Learns Physical Spectral Structure}}
\begin{figure}[t]
\centering
\includegraphics[width=\columnwidth]{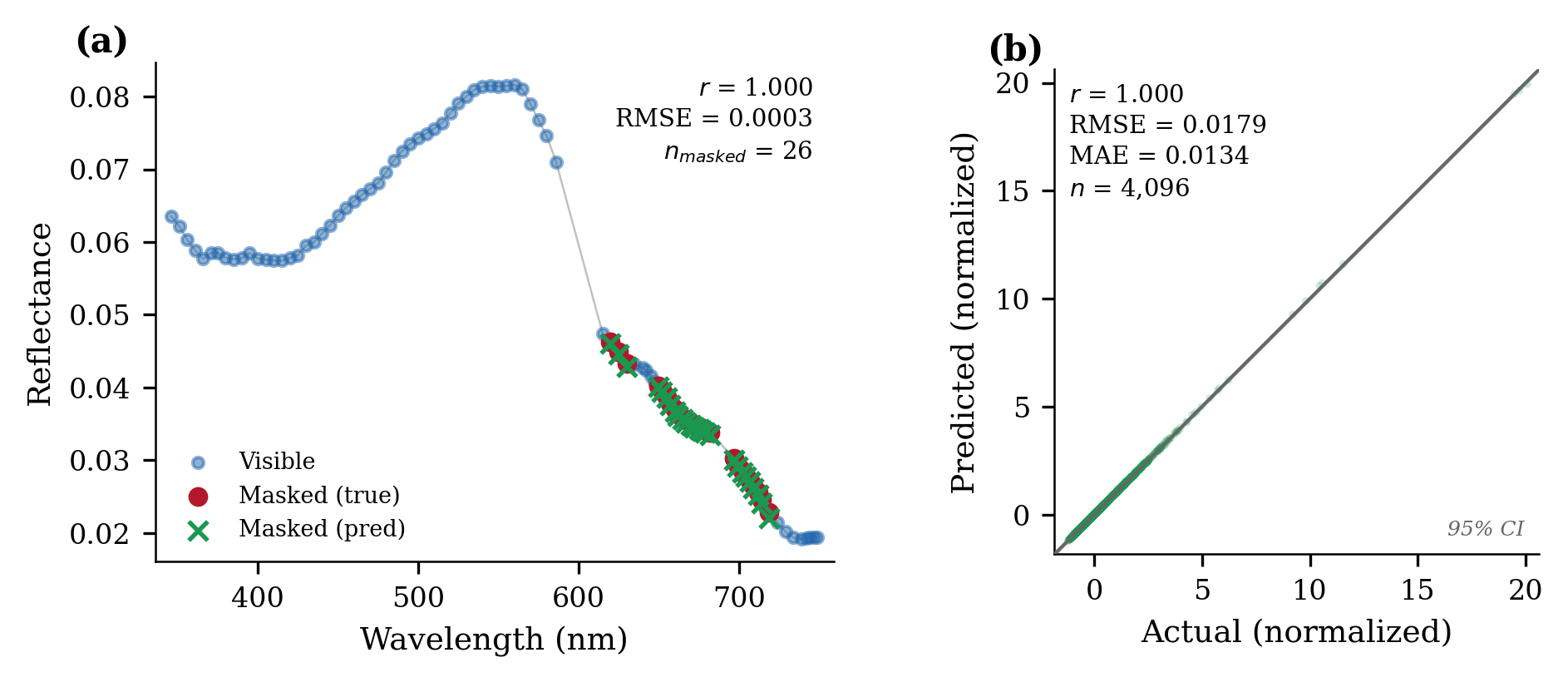}
\vspace{-7mm}
\caption{SSL pretraining validation. (a) Single-sample spectral reconstruction showing 
visible context bands (blue), true masked values (red), and predicted values (green) 
across diagnostic wavelengths (615--720\,nm). (b) Aggregate reconstruction accuracy 
across 4,096 samples ($r{=}1.000$, RMSE${=}0.018$). Shaded region: 95\% CI.}
\label{fig:ssl}
\end{figure}

SpecTM reconstructs masked diagnostic bands with near-perfect accuracy ($r{=}0.999$ on held-out validation split; Figure~\ref{fig:ssl}), despite the masking of diagnostic bands (620, 665, and 709\,nm). This demonstrates that the model learned the spectral covariance structure rather than approximating direct arithmetic computation. To contextualize reconstruction quality, we benchmarked against baseline interpolation methods: linear ($r{=}0.92$) and cubic spline ($r{=}0.96$) exhibit significant residual error in comparison to our architecture that achieves near-perfect correlation ($r{=}0.999$). This affirms that learned representations capture structural dependencies that exceed the capabilities of standard interpolation.

\subsection{\textbf{SSL Improves Microcystin Prediction}}
\label{sec:experiments}

\noindent Our pretrained architecture achieved $R^{2}$ of 0.695 for current-week prediction and 0.620 for 8-day-ahead prediction, surpassing an exhaustive benchmark of 26,208 baseline configurations (Figure~\ref{fig:combined_v3}). The pronounced performance gain in 8-day-ahead prediction (+99\% compared to +34\% for current-week) demonstrates the efficacy of temporal SSL in learning complex bloom dynamics. This 8-day lead time provides actionable warning for water managers.
\subsection{\textbf{Targeted Masking Outperforms Random Masking}}
\noindent Here, we isolate the effects of SSL pretraining via targeted ablations. We define ablation configurations as follows: 
\textit{SSL + context} uses pretrained encoder with spectral-only input; 
\textit{SSL + all features} adds meteorological auxiliary features; 
\textit{AUX-only} uses auxiliary features without the encoder; 
\textit{Random-init} replaces pretrained weights with random initialization. For random masking ablation, we matched the masking ratio (28 bands, 23\%) 
and applied contiguous masking within spectral regions to ensure fair comparison. Under identical initialization, targeted masking outperforms random masking by 0.037 $R^2$ (Figure~\ref{fig:combined_v3}), validating that domain-informed band selection improves learned representations.  The pretrained SSL features (without auxiliary) provide  increase of $R^{2}$ 0.180 relative to random initialization. This substantiates the hypothesis that SSL serves as a mechanism for distilling latent physical priors into the learned representations. The concatenation of physics-derived auxiliary features enhances performance across all configurations. The proximity of the AUX-only baseline (0.624) to the SSL + all features configuration (0.640) suggests that auxiliary features explicitly encode relationships that SSL internalizes implicitly.

\subsection{\textbf{Label Efficiency Under Scarcity}}

\begin{figure}[t]
\centering
\includegraphics[width=\columnwidth]{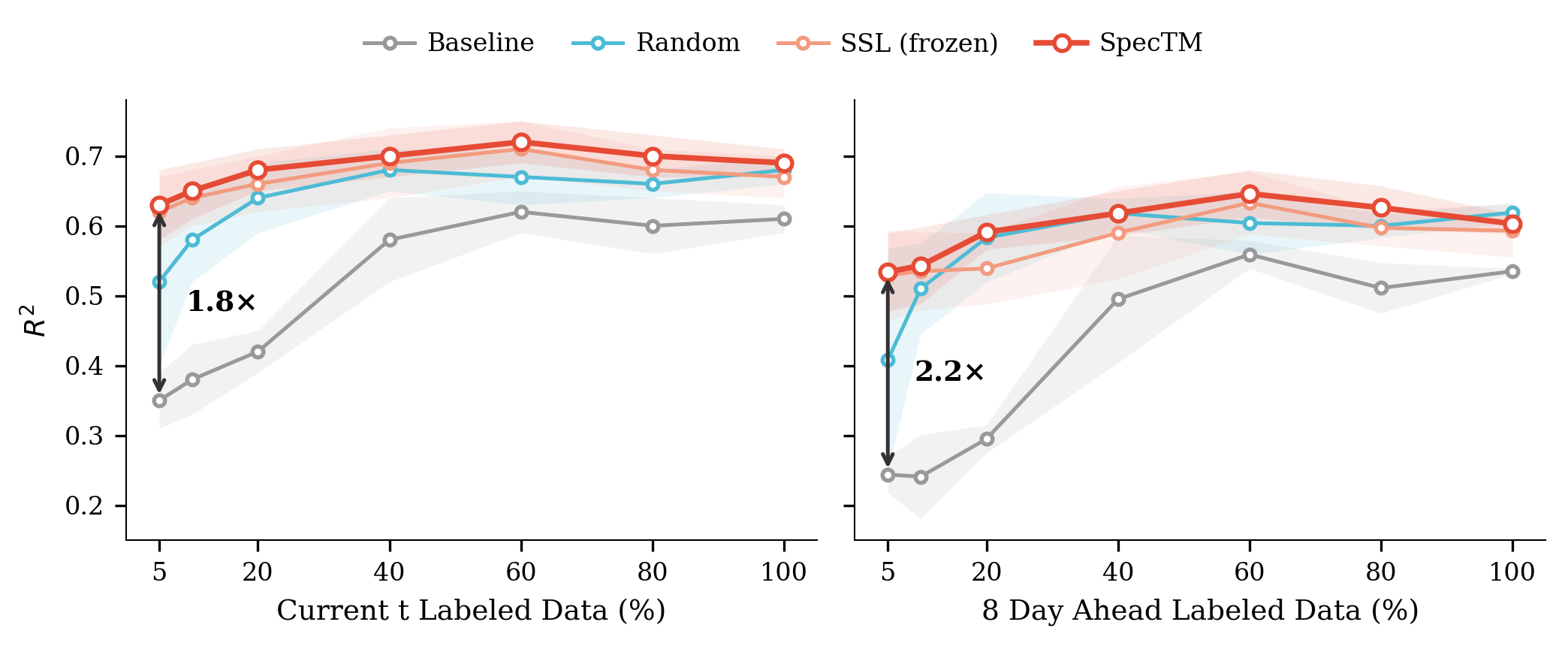}
\vspace{-7mm}
\caption{Label efficiency under data scarcity. SpecTM achieves $1.8\times$ (current) and $2.2\times$ (8-day-ahead) improvement over baseline at 5\% labeled data. Shaded regions: $\pm$1 SD across five stratified subsamples.}
\label{fig:efficiency}
\end{figure}
\noindent We evaluated label efficiency with uncertainty estimates from five stratified random subsamples per training fraction (Figure~\ref{fig:efficiency}). At extreme scarcity ($n{=}8$), SSL pretraining achieved $2.2\times$ improvement over the AUX-only baseline despite high variance inherent to such a small environmental sample size. These results substantiate that pretrained representations provide domain-informed inductive bias that persists even under minimal supervision. With increasing labeled data, the performance disparity diminishes as both methodologies converge toward a common asymptotic limit; however, SSL maintains a persistent performance advantage across all data fractions.

\section{Discussion}
\label{sec:discussion}

Targeted masking bridges the trustworthiness gap in spectral foundation models. Our pretraining objective enforces physical consistency over stochastic noise or spurious correlations to encode spectral features of intrinsic interpretability. The ablation experiments substantiate that domain-informed masking outperforms random masking. From this, we glean that the model transcends dataset-specific patterns. We find that SpecTM learns similar spectral relationships as known physics features, but through different mechanisms. This is because spectral indices are derived from the same raw spectral bands learned by SpecTM, which makes additional feature information redundant. We posit that the efficacy of SpecTM is therefore paramount in domains where such derived features are absent, such as for applications (e.g., wildfire monitoring) or target variables (e.g., nutrient concentrations) that defy analytical formulation. Our framework extends beyond aquatic applications and establishes a principled approach to encoding domain knowledge for foundation models. Diverse spectroscopic disciplines with established spectral indices, such as vegetation indices (NDVI) in agriculture and mineral absorption indices (MAI) in geology, can tailor targeted masking strategies to encode intrinsic features.

Limitation and Future Work: In this study, our experiments are confined to the water quality domain, with future work seeking to expand this scope to multi-domain applications.

\section{Conclusion}
We introduced \textit{Targeted Masking} based on a principle of physics-informed self-supervised learning, designed to mask spectral bands with known sensitivities to the target variable. In this work, we presented a representative example of cyanotoxin (\textit{microcystin} concentration) prediction as a proof of concept, where we masked phycocyanin, chlorophyll-a, and red-edge-sensitive bands to elucidate spectral covariance for bloom dynamics. We evaluated our approach with detailed experiments and our results demonstrate that this domain-informed learner strongly outperforms traditional baselines and maintains robust performance even with limited labels. As target-sensitive bands are domain-defined, we posit that this principle extends to hyperspectral applications beyond aquatic monitoring.

\bibliographystyle{IEEEtran}

\end{document}